\documentclass{article}

\usepackage{PRIMEarxiv}

\usepackage[utf8]{inputenc}
\usepackage[T1]{fontenc}
\usepackage{booktabs}
\usepackage{hyperref}
\usepackage{url}

\usepackage{booktabs}
\usepackage{amsfonts}
\usepackage{nicefrac}
\usepackage{microtype}
\usepackage{fancyhdr}
\usepackage{xcolor}
\usepackage{graphicx}
\usepackage{subcaption}

\graphicspath{{media/}}

\title{ARCADE: A City-Scale Corpus for Fine-Grained Arabic Dialect Tagging}

\author{
    \mdseries Omer Nacar\textsuperscript{1},
    Serry Sibaee\textsuperscript{2},
    Adel Ammar\textsuperscript{2}\textsuperscript{*},
    Yasser Alhabashi\textsuperscript{2}, \\
    Nadia Samer Sibai\textsuperscript{2},
    Yara Farouk Ahmed\textsuperscript{2},
    Ahmed Saud Alqusaiyer\textsuperscript{2},
    Sulieman Mahmoud AlMahmoud\textsuperscript{2}, \\
    Abdulrhman Mamdoh Mukhaniq\textsuperscript{2},
    Lubaba Raed\textsuperscript{2},
    Sulaiman Mohammed Alatwah\textsuperscript{2},
    Waad Nasser Alqahtani\textsuperscript{2}, \\
    Yousif Abdulmajeed Alnasser\textsuperscript{2},
    Mohamed Aziz Khadraoui\textsuperscript{3},
    Wadii Boulila\textsuperscript{2} \\
    \\
    \textsuperscript{1}Tuwaiq Academy, Riyadh 13415, Saudi Arabia \\
    \textsuperscript{2}Robotics and Internet-of-Things Laboratory, Prince Sultan University, Riyadh, Saudi Arabia \\
    \textsuperscript{3} Higher School of Communication of Tunis (SUP'COM), Ariana 2083, Tunisia \\
    \textsuperscript{*}Corresponding author: \texttt{aaammar@psu.edu.sa}
}

\begin{document}

\maketitle

\begin{abstract}
The Arabic language is characterized by a rich tapestry of regional dialects that differ
substantially in phonetics and lexicon, reflecting the geographic and cultural
diversity of its speakers.  Despite the availability of many multi-dialect datasets, mapping speech to fine-grained dialect sources, such as cities, remains underexplored.  We present ARCADE (Arabic Radio Corpus for Audio Dialect Evaluation), the first Arabic speech dataset designed explicitly with city-level dialect granularity. The corpus comprises Arabic radio speech collected from streaming services across the Arab world.  Our data pipeline captures 30‑second segments from verified radio streams, encompassing both Modern Standard Arabic (MSA) and diverse dialectal speech. To ensure reliability, each clip was annotated by one to three native Arabic reviewers who assigned rich metadata, including emotion, speech type, dialect category, and a validity flag for dialect identification tasks. The resulting corpus comprises 6,907 annotations and 3,790 unique audio segments spanning 58 cities across 19 countries. These fine‑grained annotations enable robust multi‑task learning, serving as a benchmark for city-level dialect tagging. We detail the data collection methodology, assess audio quality, and provide a comprehensive analysis of label distributions.
\end{abstract}
% ## Keep vs Skip Analysis
% - Keep Annotations: 4,539 (65.7%)
% - Skip Annotations: 2,366 (34.3%)

\section{Introduction}

The Arabic language exhibits substantial variation across its many dialects, which differ not only lexically and morphologically but also phonologically. While Classical Arabic and Modern Standard Arabic (MSA) provide a common formal written standard and appear in some formal media, everyday speech across all Arab countries is dialectal. Dialectal Arabic (DA) diverges from MSA and Classical Arabic in pronunciation, vocabulary, and grammar, reflecting the speaker’s geographic origin \cite{sibaee2025shamimtsyrianarabicdialect}. Prior work on the Integrated Arabic Dialect Dataset (IADD) notes that DA significantly differs from Classical Arabic and MSA in that it reflects the geographic location of the speaker \cite{zahir2021iadd}. Accordingly, we built a corpus that includes fine-grained city-level recordings to support research on dialectal variation, sociolinguistics, and culturally-aligned speech models \cite{nacar2025towards}.

Prior studies in Arabic dialect identification typically frame classification at country or large-region scales, and their corpora and evaluations follow suit. This setup offers advanced baseline performance and cross-system comparability. Yet, it provides limited support for city-level labeling, hinders analyses of urban vs. rural speech, and constrains the development of applications that benefit from finer-grained dialect signals. 

The Nuanced Arabic Dialect Identification (NADI) shared tasks provide benchmark data for assigning utterances to country‑level dialect labels and have advanced the state of the art in dialect identification \cite{abdulmageed2024nadi}. For example, the 2024 edition introduced multi‑label and multi‑level dialect identification tasks yet achieved only moderate performance \cite{abdulmageed2024nadi}. Most recently, NADI 2025 shifted focus from text to speech processing, establishing a unified benchmark for spoken dialect identification, automatic speech recognition (ASR), and diacritic restoration \cite{talafha2025nadi}. While the best identification systems achieved 79.8\% accuracy, the results underscore the persisting challenges of processing multidialectal spoken Arabic.
At the audio level, the ADI17 dataset collected roughly 3,000 hours of
YouTube speech covering 17 Arabic countries for dialect
identification \cite{shon2020adi17}. These datasets provide coarse
geographical labels (typically at the country or region level), are limited to high‑level dialect classes, and offer little information about intra‑country variation.

In contrast, this work presents a corpus of short radio segments annotated with city-level labels and detailed metadata, supporting fine-grained dialect modeling and evaluation.  Our contributions are
threefold:

\begin{itemize}
\item We design and implement a data pipeline that records radio streams from
  numerous Arab cities, filters for monologue segments, and annotates clips with emotion, speech type, dialect category, and quality labels.
\item We analyze the resulting corpus, showing distributions of emotion,
  speech type, keep/skip decisions, dialect categories, and confidence labels.  The dataset comprises 58 cities across 19 Arab countries, with at least 10 recordings per city, providing broad geographic coverage.
\item We outline practical use cases and key considerations for fine-grained modeling. Benchmarking and modeling are left for future work.
\end{itemize}

By releasing both the data and a transparent collection and annotation protocol, we aim to catalyze research on fine-grained dialect attribution from audio. The corpus enables city-level labeling, dialect-provenance modeling, robustness studies across channel and domain shifts, and transfer to related media such as telephony and online video. The reusable protocol supports future expansions, community contributions, and reproducible benchmarks for fair modeling comparisons.

\section{Related Works}

This section surveys the progression of Arabic dialect identification research, spanning both textual and acoustic perspectives. Early work relied primarily on text-based corpora, from regional datasets to later multi-label benchmarks, revealing systematic divergences between Dialectal Arabic and Modern Standard Arabic. More recent efforts have shifted toward spoken data, with large-scale broadcast and web-derived speech corpora enabling advances in acoustic modeling. Despite this progress, existing resources remain limited by coarse dialect labels and uneven annotation quality, motivating the need for datasets with finer-grained and more reliable dialect annotations.

\subsection{Textual Resources for Dialect Identification}

Research on Arabic dialect identification has historically been driven by textual resources. Early corpora such as PADIC \cite{meftouh-etal-2015-machine}, SHAMI \cite{nayouf-etal-2023-nabra, sibaee2025shamimtsyrianarabicdialect}, and TSAC \cite{medhaffar-etal-2017-sentiment} focused on dialect classification from text. These corpora were later consolidated into larger benchmarks such as the Integrated Arabic Dialect Dataset (IADD) \cite{zahir2021iadd}, which merges subsets of five public datasets covering Levantine, Tunisian, Egyptian, Maghrebi, Iraqi, and Gulf dialects. This work highlights that Dialectal Arabic (DA) systematically diverges from Modern Standard Arabic (MSA) and Classical Arabic (CA) through distinct regional variations, which are critical for identification. Recent efforts have further advanced Arabic NLP by developing general-purpose text embeddings to better capture semantic and dialectal nuances \cite{nacar2025gate}; however, applying these textual insights to the acoustic domain remains a distinct challenge.

Shared tasks have also played a pivotal role in advancing textual dialect identification. The Nuanced Arabic Dialect Identification (NADI) series initially operated at the text level, providing country-level labels for tweets. For instance, the 2024 edition introduced multi-label country identification and dialect intensity prediction tasks \cite{abdulmageed2024nadi}. Despite these advances, text-based identification remains distinct from the challenges posed by acoustic variation in spoken dialects.

\subsection{Spoken Arabic Corpora and Benchmarks}

The development of robust Automatic Speech Recognition (ASR) and spoken dialect identification systems has relied heavily on large-scale broadcast and web-scraped corpora. Early efforts, such as the CALLHOME Egyptian Arabic corpus \cite{karins2002callhome}, provided foundational resources for conversational telephone speech.

More recently, the Multi-Genre Broadcast (MGB) challenge series has established major benchmarks for the field. While MGB-1 \cite{bell2015mgb} focused on English, MGB-2 \cite{7846277} introduced a massive Arabic resource comprising over 1,200 hours of Al Jazeera broadcasts, covering MSA and four broad dialect regions (Egyptian, Gulf, Levantine, and North African), and MGB-3 \cite{8268952} shifted focus to ''in-the-wild'' dialectal speech, utilizing multi-genre YouTube videos primarily for Egyptian ASR.

To address relatively fine-grained identification, the ADI17 corpus \cite{shon2020adi17} collected approximately 3,000 hours of YouTube audio from 17 countries. This dataset was subsequently used in the MGB-5 challenge \cite{ali2019mgb}, which targeted country-level dialect identification. Similarly, the QCRI Arabic Speech Recognition (QASR) dataset \cite{mubarak-etal-2021-qasr} provides 2,000 hours of broadcast speech with linguistic segmentation and named entity annotations.

Table \ref{tab:resources} provides a comparative summary of these primary resources. Despite recent advances, most existing corpora provide coarse-grained dialect labels at the country or regional level (e.g., Egyptian, Levantine). Furthermore, there is a significant distinction in label reliability. Massive collections like MGB-5 \cite{ali2019mgb} and ADI17 \cite{shon2020adi17} rely on weak supervision, where dialect labels are inferred from channel metadata, inherently introducing noise into the training data \cite{ali2019mgb}. Similarly, QASR relies on light supervision for transcription and automatic classification for dialect analysis, with only a small subset manually annotated for country-level evaluation \cite{mubarak-etal-2021-qasr}.

While our dataset is not the largest in terms of duration or number of recordings, its primary value added lies in the combination of granularity, breadth, and label quality. Our work differs by providing fine-grained sub-regional dialect labels across the largest number of Arab countries (19), rigorously verified by 11 trained native Arab annotators, with one to three annotations per audio segment. This comprehensive annotation process captures detailed metadata, including speaker emotion, speech type (e.g., single/multiple speakers, Quran recitation, music), content validity (keep/skip), dialect category (MSA, dialect, mixed), and annotator confidence, addressing the critical gap in geographically precise and reliable dialect identification resources.

% For code-switching phenomena, Samih et al. \cite{samih-etal-2016-multilingual} developed resources capturing the interplay between dialectal Arabic and other languages. The Corpus of Regional Arabic \cite{habash-etal-2012-conventional} systematically documents phonological and morphological variation across regions.

% Recent work has emphasized the development of universal and multilingual approaches. Wang et al. \cite{wang2024universal} note that Arabic ASR research faces unique challenges due to rich morphology and dialectal variation; developing models that generalise across dialects requires extensive multi-dialect data. Their work on universal Arabic ASR systems demonstrates the importance of large-scale, diverse training data. Similarly, Alghamdi et al. \cite{article} present large-scale Arabic ASR resources that span multiple dialects and speaking conditions. Baevski et al. \cite{10.5555/3495724.3496768} introduced self-supervised learning approaches that have proven effective for low-resource Arabic dialects, while Abdelali et al. \cite{9780142} explored speech translation across Arabic varieties.

\begin{table}[h]
\centering
\caption{Comparison of ARCADE with major existing Arabic speech resources. Note the distinction between resources relying on noisy/weak labels versus those with verified annotations.}
\label{tab:resources}
\resizebox{\columnwidth}{!}{%
\begin{tabular}{@{}lcclclc@{}}
\toprule
\textbf{Resource} & \textbf{Size (Clips / Duration)} & \textbf{Type} & \textbf{Key Focus} & \textbf{Granularity} & \textbf{Labeling} & \textbf{Countries} \\ \midrule
ADI17 \cite{shon2020adi17} & ${\sim}$1M clips (3,000 Hours) & 17 Dialects & YouTube (Fine-grained) & Country-level & Weak (Channel-based) & 17 \\
MGB-2 \cite{7846277} & ${\sim}$550k clips (1,200 Hours) & MSA + 4 Regions & Broadcast (Al Jazeera) & Regional & Lightly Supervised & -- \\
MGB-3 \cite{8268952} & 13,825 clips (16 Hours) & EGY + 5 Dialects & YouTube (Multi-genre) & Regional & Manually Verified & 5 \\
MGB-5 \cite{ali2019mgb} & ${\sim}$1.06M clips (3,000 Hours) & 17 Dialects & YouTube (Fine-grained) & Country-level & Weak (Channel-based) & 17 \\
QASR \cite{mubarak-etal-2021-qasr} & 1.6M clips (2,000 Hours) & MSA + 4 Regions & Broadcast (Al Jazeera) & Regional & Lightly Supervised & -- \\
CALLHOME \cite{karins2002callhome} & 120 calls (${\sim}$60 Hours) & Egyptian & Telephone speech & Dialect-specific & Manually Verified & 1 \\
\textbf{ARCADE (Ours)} &  3,790 clips (31.6 Hours) & MSA + Dialects & Radio Streams & \textbf{City-level} & \textbf{Manual (multiple annotations)} & \textbf{19} \\ \bottomrule
\end{tabular}%
}
\end{table}

\section{Dataset Collection and Annotation}

\subsection{Recording Pipeline}

\begin{figure}[t]
  \centering
  \includegraphics[width=\textwidth]{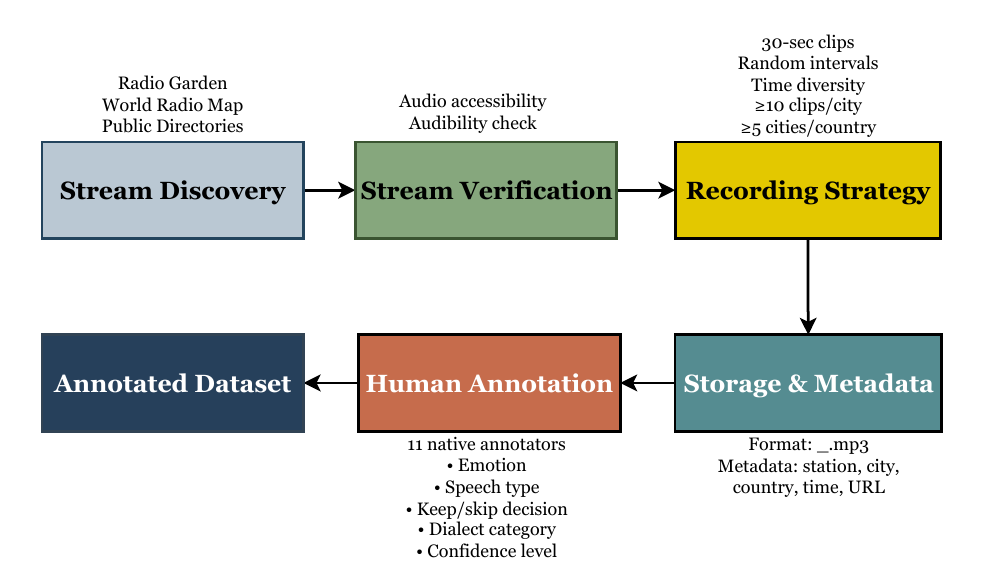}
  \caption{The main pipeline for building the ARCADE dataset.}
  \label{fig:pipeline}
\end{figure}

We developed a pipeline that automatically discovers and records radio streams
across the Arab world (Figure \ref{fig:pipeline}).  Streaming sources include Radio Garden\footnote{\url{https://radio.garden/}}, World
Radio Map\footnote{\url{https://worldradiomap.com/}}, and other public radio directories. Each stream is validated to confirm that the audio is both accessible and of sufficient quality for analysis. Once verified, the system records 30-second clips at random intervals. This duration was deliberately chosen to balance robustness and homogeneity, as prior work has shown that 30-second segments yield strong performance for Arabic dialect identification (exceeding 81\% accuracy) \cite{Biadsy}, while also increasing the likelihood that each recording contains a single speaker and a consistent speech style. We aim for monologue-like speech and ensure a diverse range of times of day to capture different programmes and presenters (news, variety shows, cultural segments, and similar formats).  Clips are stored with structured filenames
that encode city and timestamp (\texttt{<city>\_<timestamp>.mp3}).  Metadata
captured for each clip includes the station name, city, country, start time, and stream URL.

To broaden coverage, we set a minimum of 10 kept recordings per city and aimed to sample at least 5 cities per country; however, this target could not be met consistently because several countries lack 5 identifiable radio stations from different cities that publicly broadcast on the Internet. Cities are chosen based on
population, media prominence, and geographic spread.  Dataset was annotated by 11 trained native Arab annotators who identified the speaker’s emotion (e.g., neutral, happiness, or anger), labeled the speech type (regular single speaker, multiple speakers, no speech/music, or Quran recitation), judged whether the clip should be kept or skipped, and determined the dialect category (MSA, dialect, mixed, or not applicable), and provided a confidence level (sure, unsure, or no idea) to verify the classification, ensuring that non-local speakers appearing on local radio stations are flagged. 

\begin{figure}[ht]
  \centering
  \includegraphics[width=\textwidth]{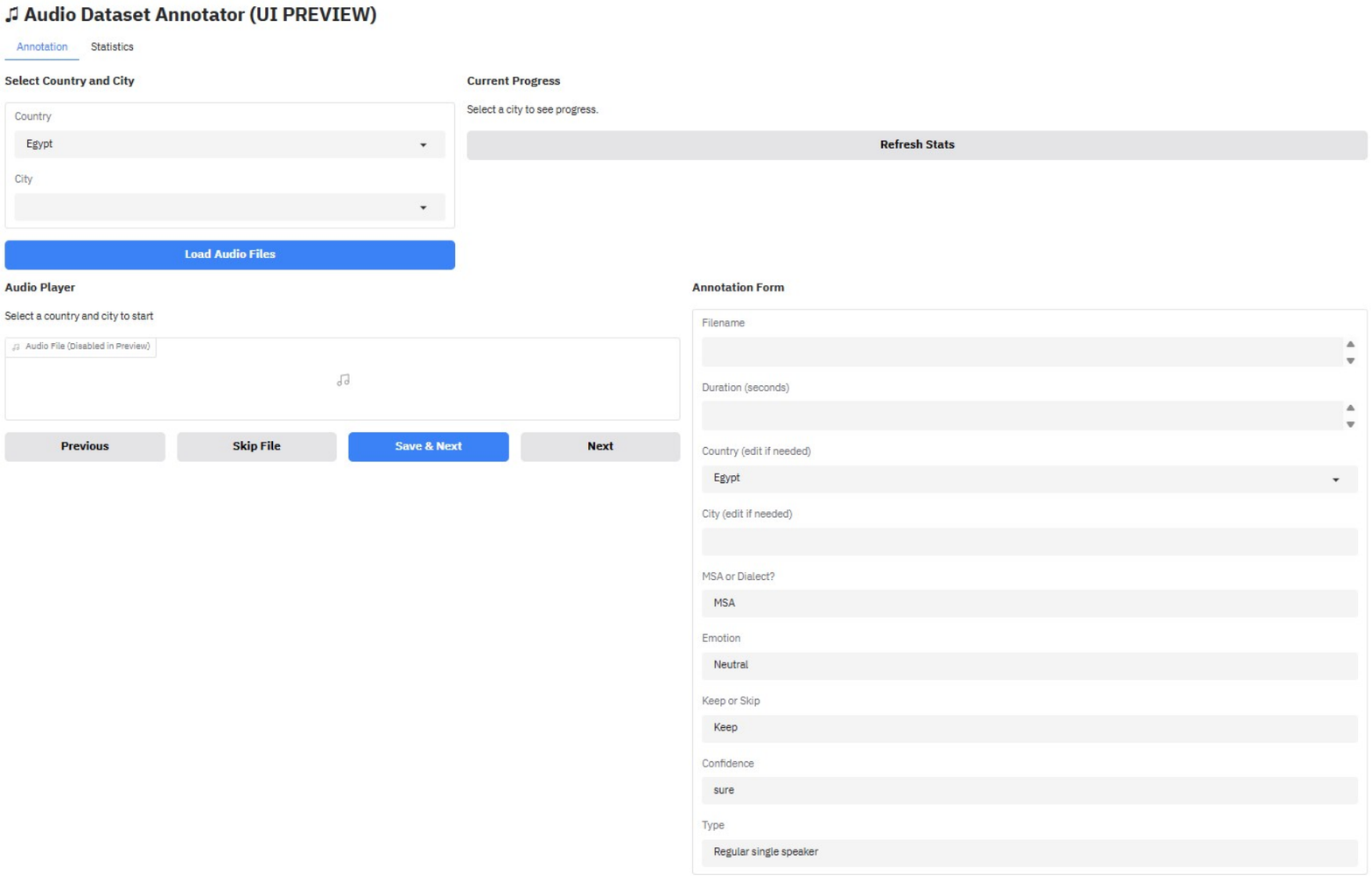}
  \caption{Screenshot of the custom Gradio-based annotation interface used by native speakers to label the dataset.}
  \label{fig:gradio_interface}
\end{figure}

To support efficient annotation, we developed a custom web-based interface built on the Gradio framework (Figure \ref{fig:gradio_interface}). The interface enables annotators to play audio segments while concurrently assigning labels for emotion, speech type, dialect category, and confidence level through standardized dropdown menus. It integrates navigation controls for sequential labeling or skipping and features a real-time statistics dashboard to monitor individual and global progress, ensuring data consistency across the annotation team.

%  Overall Statistics
% Total annotations: 6,907
% Unique files: 3,602
% Unique countries: 19
% Unique cities: 60
% Unique annotators: 11

\subsection{Data Overview}
\begin{figure}[t]
  \centering
  \includegraphics[width=0.9\textwidth]{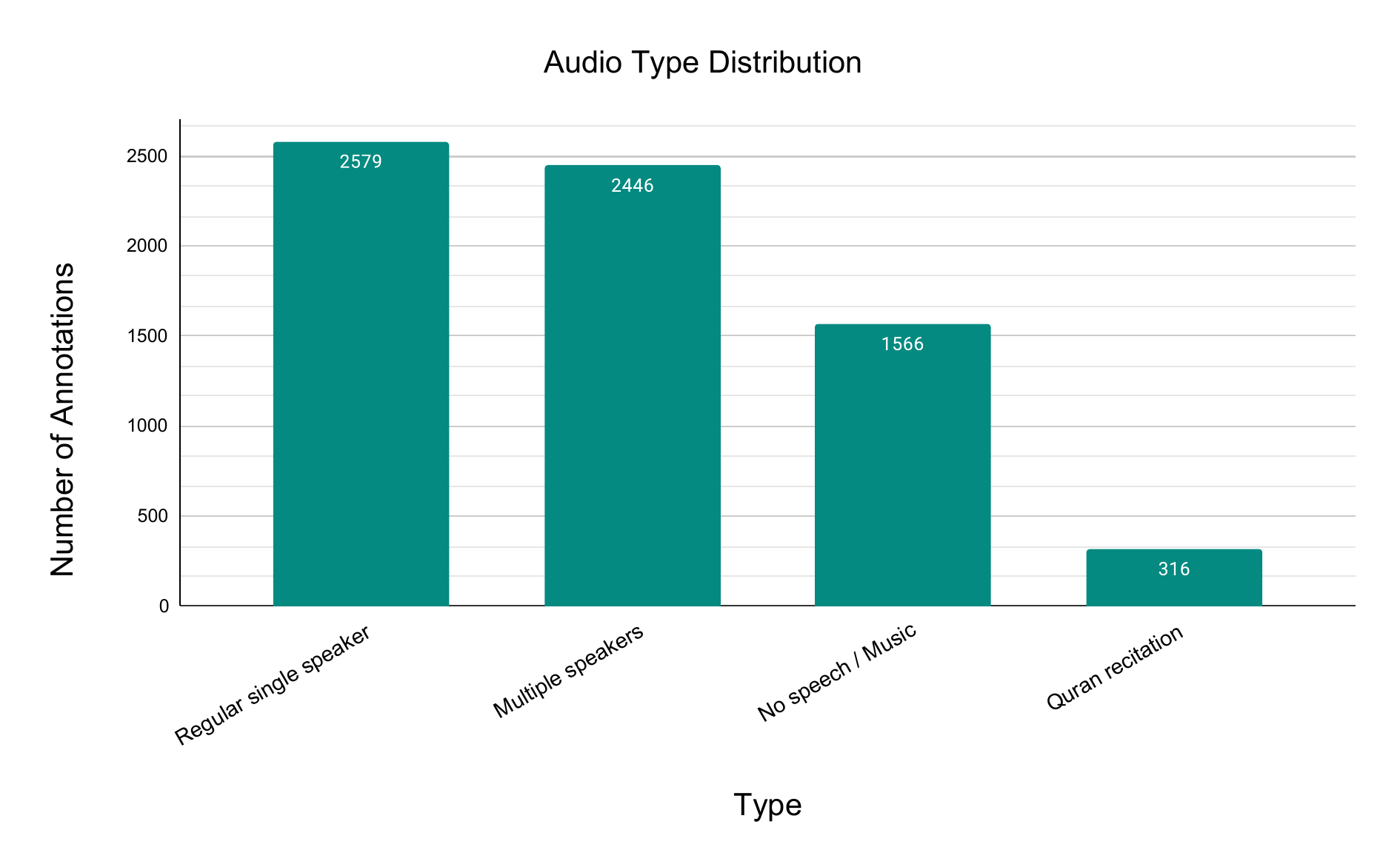}
  \caption{Distribution of audio types across the annotated corpus. The dataset primarily consists of regular single-speaker recordings and multiple-speaker segments, with a smaller subset containing music/no speech or Quranic recitation.}
  \label{fig:types}
\end{figure}

The current release comprises 6,907 annotations and 3,790 unique audio clips. Of these, 65.7\% were manually marked as "keep" to be retained for dialect identification tasks, while the remaining 34.3\% were marked as "skip" in the dataset due to the predominance of Quranic recitation, music, or crosstalk. Post-annotation analysis reveals that regular single-speaker speech constitutes 37.3\% of the segments, whereas 35.4\% feature multiple speakers; additionally, 22.7\% contain music or non-speech audio, and 4.6\% are identified as Quranic recitations (Figure~\ref{fig:types}). The latter were excluded because religious recitation adheres to the strict, standardized phonology of Classical Arabic \cite{qandos2025anplers}, which suppresses the phonotactic variations necessary for dialect identification. Besides, broadcasts frequently feature renowned transnational reciters rather than local speakers, thereby dissociating the speaker's origin from the station's location.

Table~\ref{tab:annotation_fields} describes the annotation schema used in the ARCADE dataset. Each audio clip is stored in MP3 format with its corresponding \textit{Filename} and \textit{Duration} in seconds. Geographic metadata captures both the \textit{Country} and \textit{City} of origin, enabling the fine-grained city-level dialect analysis that distinguishes ARCADE from existing resources. The core linguistic annotation indicates whether each clip contains Modern Standard Arabic or dialectal speech (\textit{MSA or Dialect}), while the \textit{Audio type} field categorizes clips as containing a regular speaker, multiple speakers, or no speech. The \textit{Emotion} field captures the emotional tone of the speaker(s). Quality control is maintained through the \textit{Keep or skip} field for filtering unsuitable samples and the annotator's self-reported \textit{Confidence} level. Full traceability is ensured via \textit{Annotator} identification and \textit{Timestamp} fields, supporting reproducibility of the annotation process.

% explaining the rows of the dataset
\begin{table}[h!]
\centering
\caption{Description of dataset annotation fields}
\label{tab:annotation_fields}
\begin{tabular}{l p{8cm}}
\hline
\textbf{Field} & \textbf{Description} \\
\hline
Audio & Audio file in mp3 format. \\
Filename & Name of the audio file. \\
Duration & Length of the audio clip in seconds. \\
Country & Country where the speaker or recording originates. \\
City & City of the speaker or recording. \\
MSA or Dialect? & Modern Standard Arabic or a specific dialect. \\
Emotion & The emotional tone expressed in the audio (See Figure~\ref{fig:emotion}). \\
Keep or skip &Retained or discarded for dialect identification tasks. \\
Confidence & Annotator's confidence level in the labeling (sure, unsure, or no idea). \\
Audio type & Category or type of speech/audio (regular speaker, multiple speakers, Quran recitation, or music). \\
Annotator & Name of the annotator who labeled the sample. \\
Timestamp & Time when the annotation was completed. \\
\hline
\end{tabular}
\end{table}

\begin{figure}[t]
  \centering
  \includegraphics[width=0.7\textwidth]{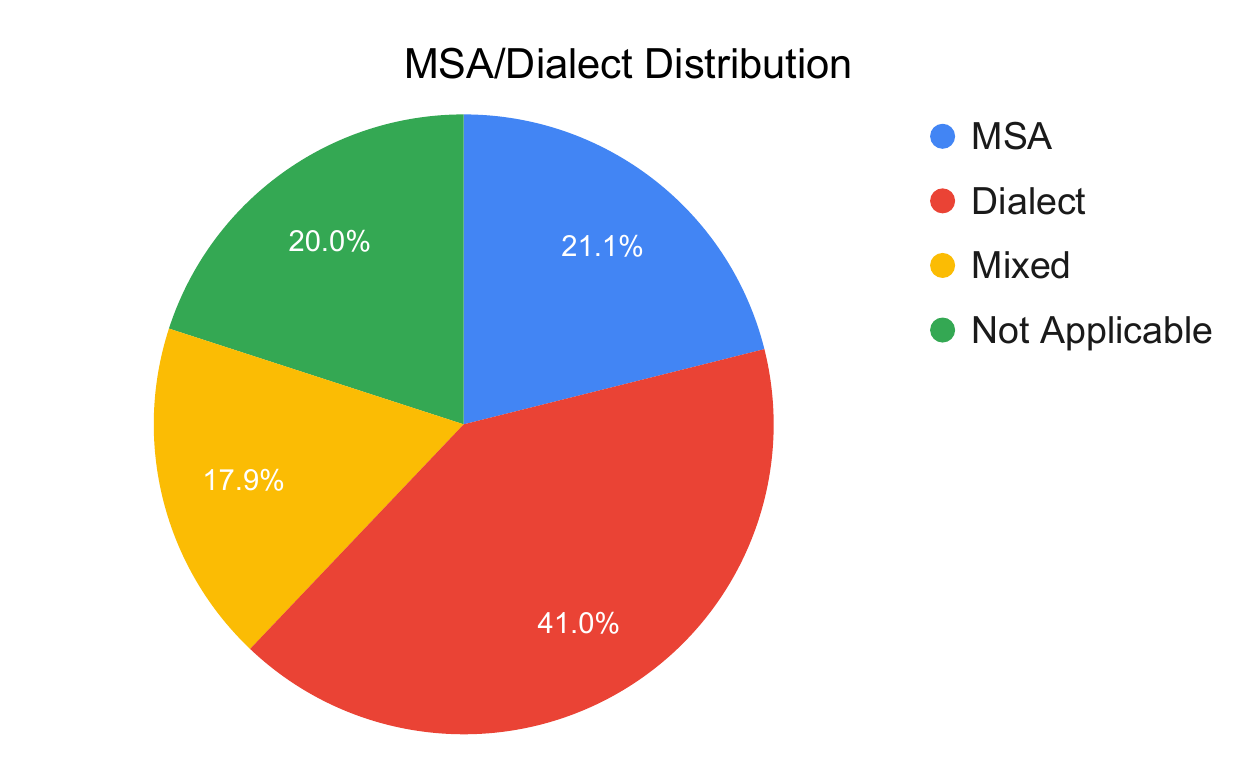}
  \caption{MSA/dialect breakdown of the collected annotated segments.  Dialectal
    speech forms the largest portion of the data, followed by MSA, then mixed
    and not applicable categories.}
  \label{fig:dialect}
\end{figure}

As shown in Figure~\ref{fig:dialect} (resulting from labeling), dialectal speech accounts for 41\% of clips, 21\% are MSA, 18\% are mixed, and 20\% are not applicable (the music or background noise overwhelms the speaker's voice). MSA clips were retained because a speaker's regional origin can often still be inferred from accentual cues even when producing standardized MSA. In fact, regional identity is preserved through phonological interference and prosodic features, resulting in distinct ``regional standards'' of MSA \cite{Biadsy,versteegh2014arabic,kalaldeh2018acoustic,abdul2016bassiouney}. 
% Most annotators reported high confidence: 91.9\% of annotations were labelled as sure'', 7.1\% as unsure'', and only 1.0\% as ``no idea'' (Figure~\ref{fig:confidence}).

% \begin{figure}[t]
%   \centering
%   \includegraphics[width=0.55\textwidth]{keep_vs_skip_distribution.png}
%   \caption{Distribution of keep vs skip decisions.  Approximately
%     two‑thirds of the recorded clips are judged suitable for modelling.}
%   \label{fig:keep}
% \end{figure}

\begin{figure}[h]
  \centering
  \includegraphics[width=0.9\textwidth]{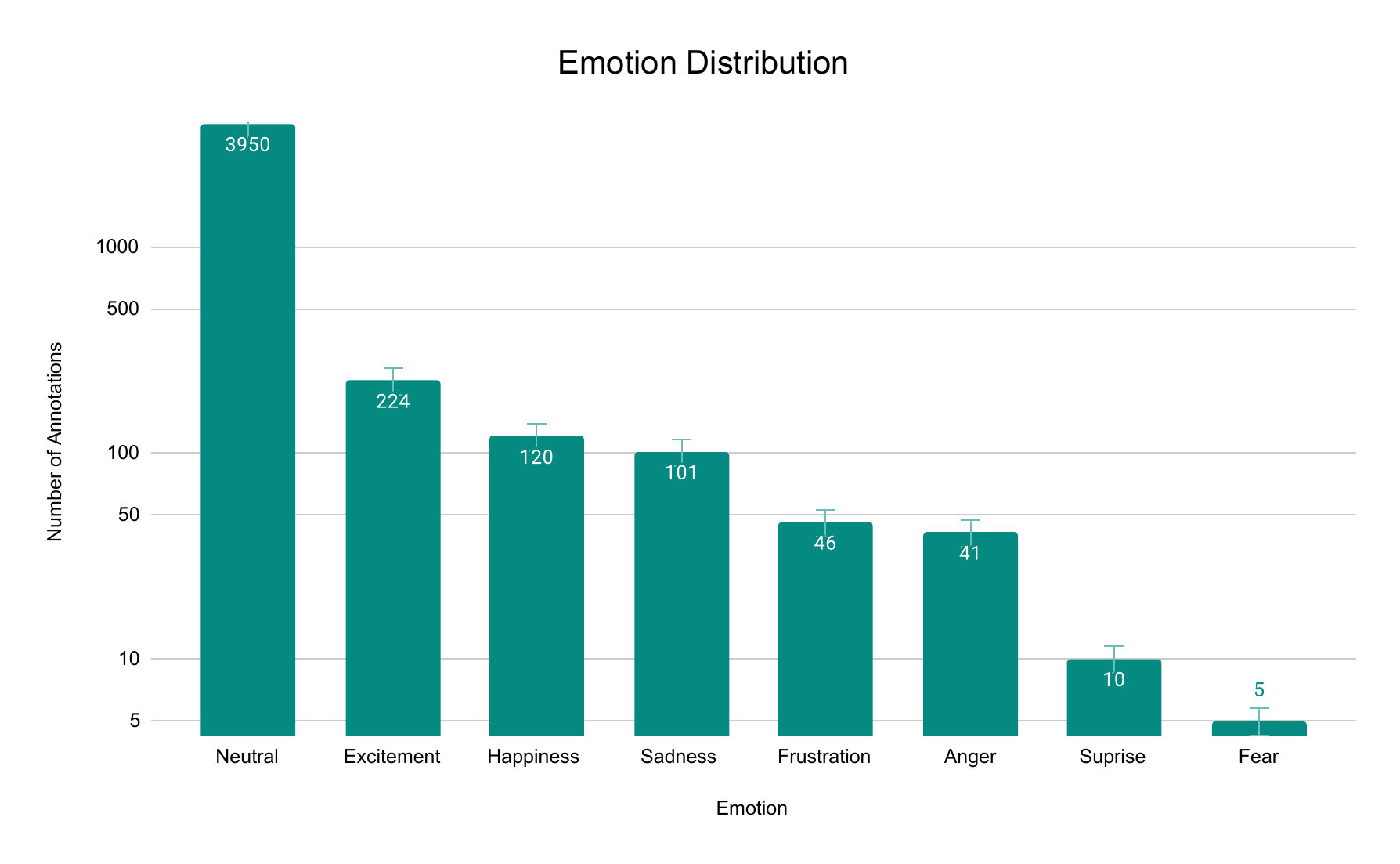}
  \caption{Emotion distribution (in log scale) in the ARCADE dataset. Emotion annotations were mainly applied to retained segments, as skipped audio clips (typically Quran recitation or music) were not meant to be labeled for emotion. Neutral accounts for the vast majority of annotations, while other emotions are scarce. Note that 42 retained recordings were missing emotion annotations and are excluded from this graph.}
  \label{fig:emotion}
\end{figure}

Figure~\ref{fig:emotion} summarizes the distribution of annotated emotions.
Neutral emotions dominate, with 87.8\% of annotations, while other
emotions such as excitement, happiness, sadness, frustration, anger,
surprise, and fear occur much less frequently. The dataset’s imbalance will inform sampling
  strategies for any downstream emotion recognition models.

% The cross‑tabulation of
% emotion and speech type (Figure~\ref{fig:crosstab}) reveals that neutral
% emotions appear across all types, whereas positive emotions such as
% excitement and happiness are more common in regular single‑speaker and
%   multiple‑speaker clips.  

% \begin{figure}[t]
%   \centering
%   \includegraphics[width=1.1\textwidth]{pdf_graphs/emotion_type_crosstab (1).pdf}
%   \caption{Cross‑tabulation of emotion versus speech type.  Neutral
%     segments span all types; excitements are concentrated in regular and
%     multiple‑speaker clips.}
%   \label{fig:crosstab}
% \end{figure}

\subsection{Geographic Coverage}

\begin{table}[h!]
\centering
\caption{Countries and cities covered in the ARCADE dataset.}
\label{tab:countries_cities}
\begin{tabular}{|l|p{10cm}|}
\hline
\textbf{Country} & \textbf{Cities} \\
\hline
Algeria & Algiers, Annaba, Batna, Blida, Biskra, Constantine, Djelfa, Oran, Setif \\
Bahrain & Manama \\
Egypt & Alexandria, Cairo \\
Iraq & Baghdad, Basra \\
Jordan & Amman, Irbid \\
Kuwait & Kuwait City\\
Lebanon & Beirut, Cheikh Taba \\
Libya & Benghazi, Tripoli \\
Morocco & Casablanca, Fes, Marrakech, Rabat, Tangier \\
Oman & Muscat, country wide \\
Palestine & Hebron, Jerusalem, Nablus, Gaza, Ramallah \\
Qatar & Doha \\
KSA & Jeddah, Makkah, Riyadh \\
Somalia & Daljir, Mogadishu, Shabelle, Somali \\
Sudan & El Obeid, Khartoum, Omdurman, Port Sudan, Wad Medani \\
Syria & Aleppo, Damascus \\
Tunisia & Tunis \\
UAE & Abu Dhabi, Ajman, Dubai, Fujairah, Sharjah \\
Yemen & Aden, Al Hodeidah, Sanaa, Taiz \\
\hline
\end{tabular}
\end{table}

\begin{figure}[t]
  \centering
  \includegraphics[width=0.95\textwidth]{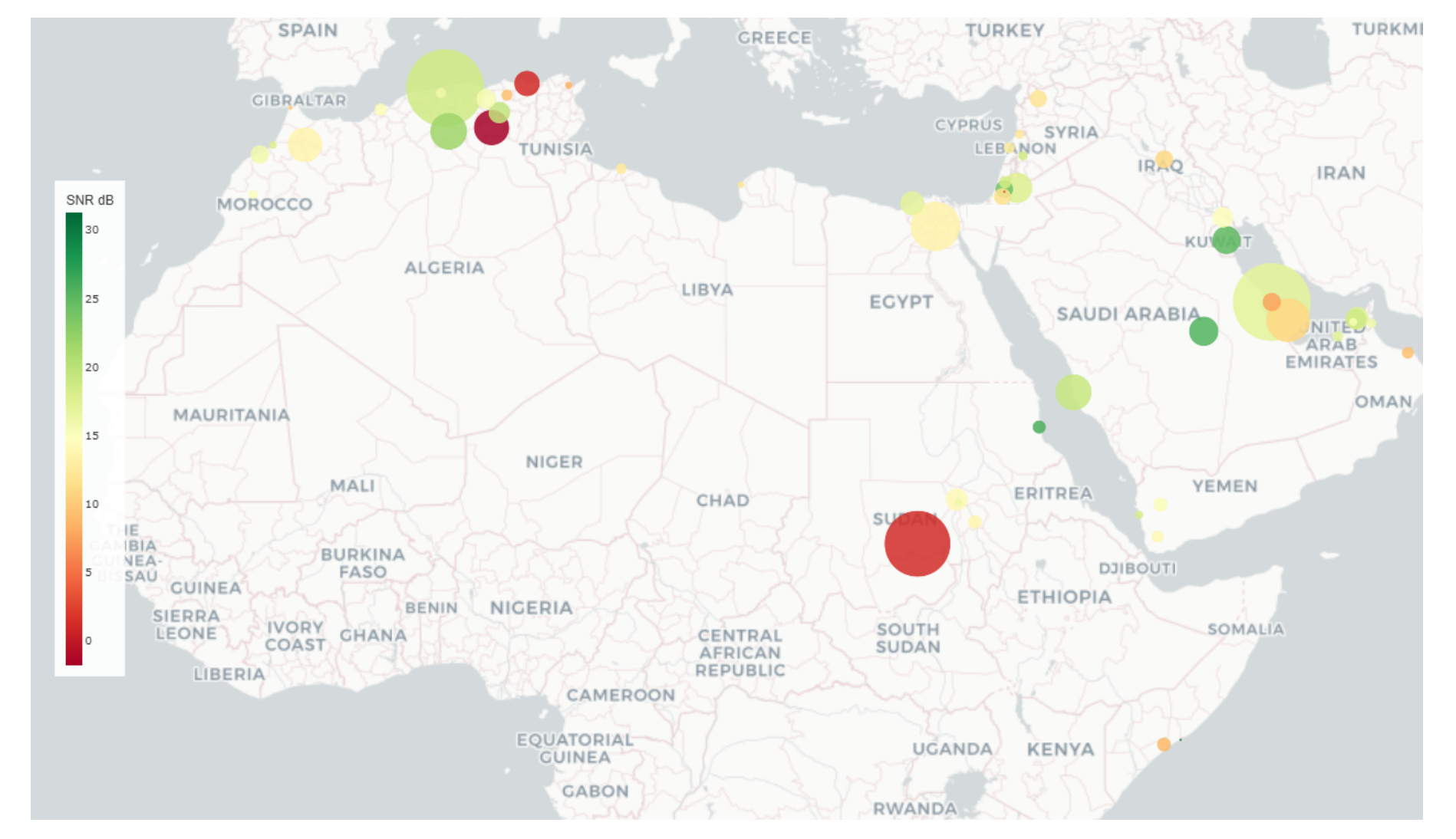}
  \caption{Spatial distribution of the dataset across the Arab world. Markers indicate city locations; the size of each circle is proportional to the volume of recordings collected, and its color indicates the mean SNR of the recordings for each city. An interactive visualization is accessible at \url{https://riotu-lab.github.io/arabic-cities-map/}.}
  \label{fig:geo}
\end{figure}

\begin{figure}[t]
  \centering
  \includegraphics[width=1\textwidth]{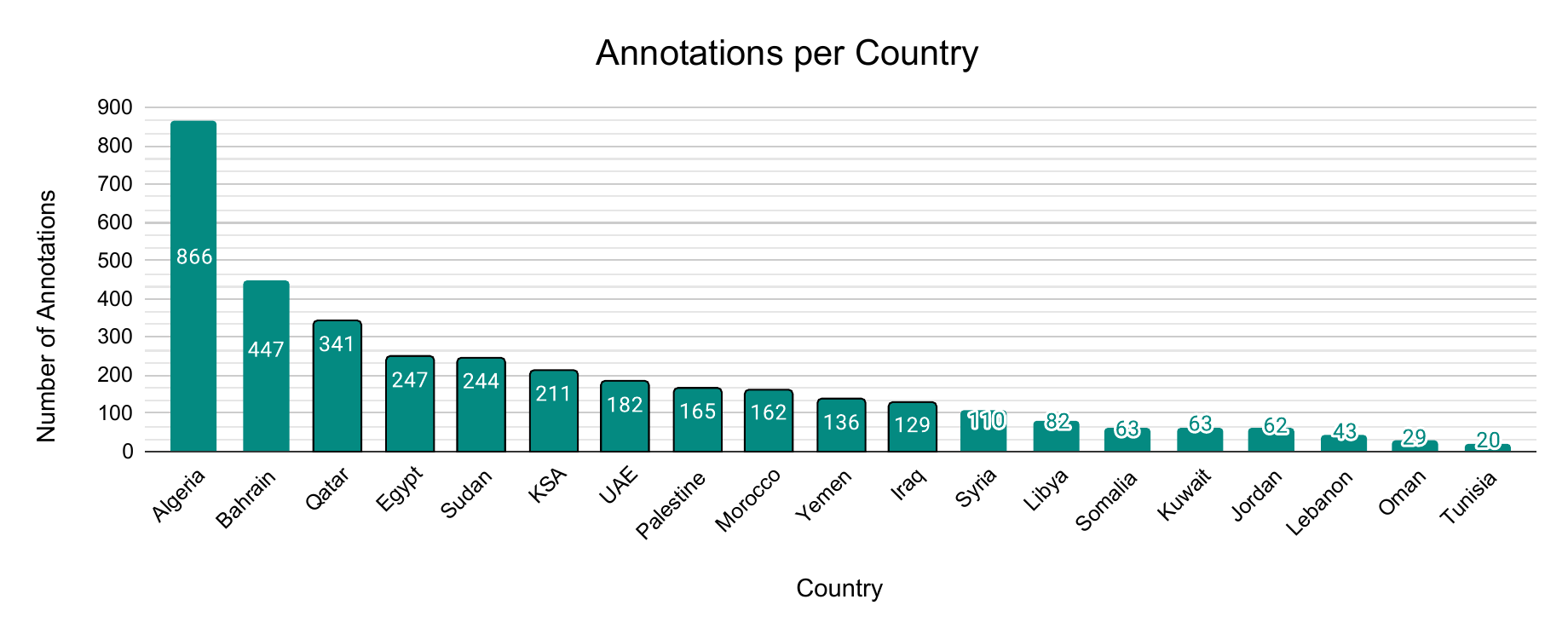}
  \caption{Distribution of annotated recordings by country. Algeria and Bahrain yield the highest volume of data (21.8\% and 14.7\%, respectively), while Oman and Tunisia yield the lowest (0.8\% and 0.6\%, respectively). 
  % The dataset covers 19 countries, with the majority contributing between 100 and 400 clips each.
  }
  \label{fig:topcountries}
\end{figure}

We \textbf{collected audio segments} from 58 cities spanning Africa, the Levant, Iraq, and the Gulf (all cities are shown in Table \ref{tab:countries_cities}). Figure~\ref{fig:geo} presents the geographic distribution of all retained radio stations. The recordings are unevenly distributed, with notable concentrations in Algeria and Bahrain, which contributed 1,509 and 1,014 clips, respectively, followed by Sudan and Qatar (Figure~\ref{fig:topcountries}). Future work may need to address this imbalance through targeted data collection or by using weighted loss functions during model training. We leave all experimental benchmarking and modeling to future research and focus here on describing the corpus.

\subsection{Technical Validation}
To ensure the reliability and utility of the ARCADE corpus, we conducted a comprehensive technical validation encompassing both annotation quality and acoustic signal characteristics. This validation addresses two complementary dimensions: first, the consistency and reliability of human annotations; and second, the acoustic quality of the recorded audio, characterized through signal-level measurements. Together, these analyses establish the corpus as a robust resource for fine-grained dialect identification research and inform potential users about data quality considerations that may influence modeling decisions.

\subsubsection{Annotation quality}
\begin{figure}[h]
    \centering
    \includegraphics[width=0.75\linewidth]{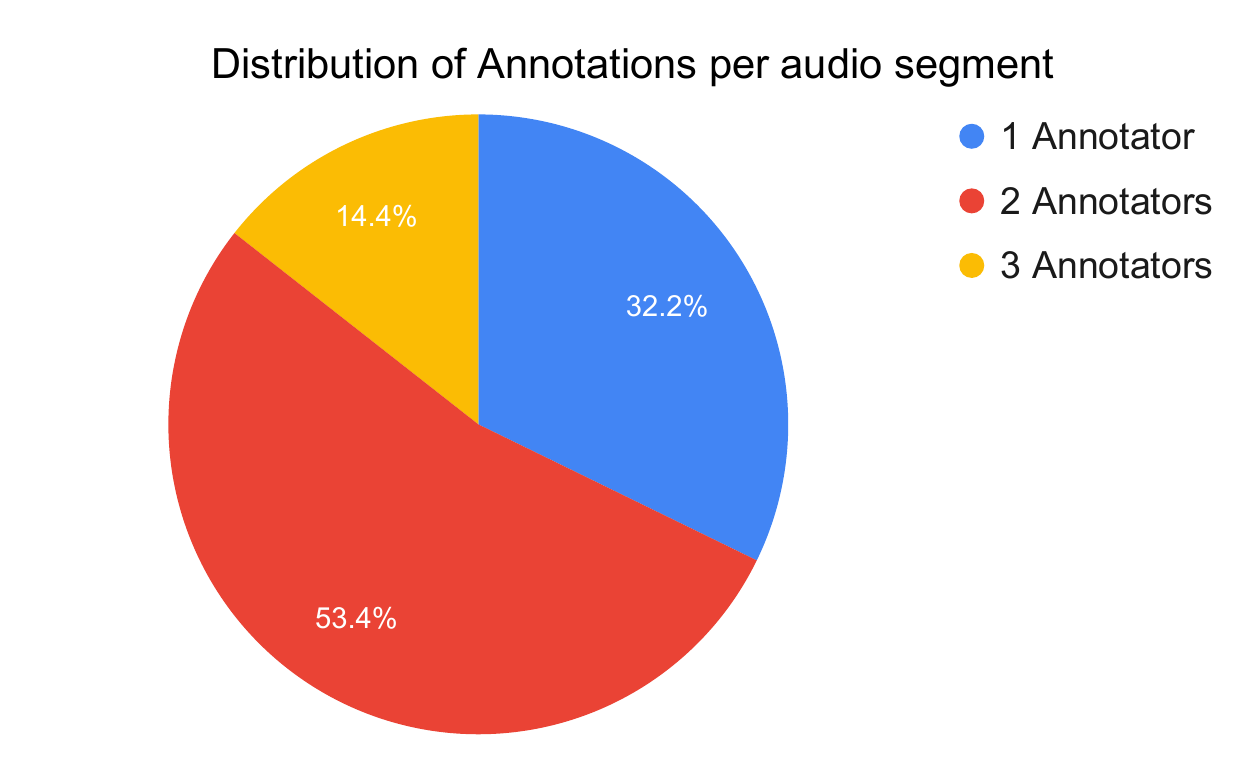}
\caption{Distribution of the number of annotators per audio segment. Over two-thirds of the corpus (67.8\%) was validated by multiple annotators (2 or 3), while 32.2\% received a single annotation pass.}    \label{fig:annotation-distribution}
\end{figure}

\begin{figure}[h]
  \centering
  \includegraphics[width=0.7\textwidth]{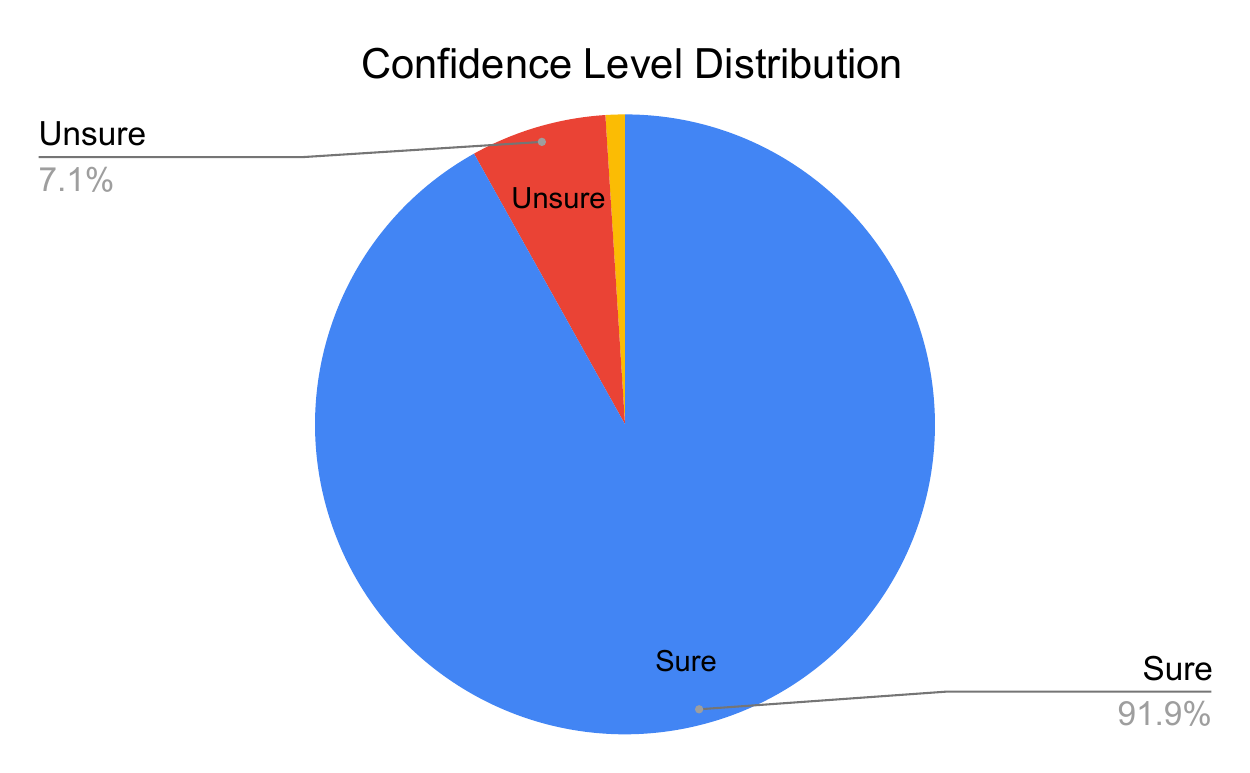}
  \caption{Annotator confidence levels.  The vast majority of labels were
    marked as “sure”, indicating reliable annotations.}
  \label{fig:confidence}
\end{figure}

Each audio clip was independently annotated by trained native Arab speakers. We initially targeted a redundancy of three annotators per segment to ensure high reliability; however, due to time constraints where some reviewers did not complete their assigned batches, this coverage varies. As shown in Figure \ref{fig:annotation-distribution}, 14.4\% of the corpus reached the full target of three reviewers and 53.4\% were validated by two, providing a robust basis for consensus for the majority of the data. The remaining 32.2\% received a single annotation pass. Annotators labeled emotion, speech type, quality decision (keep/skip), dialect category (MSA, dialectal, mixed, not applicable), and confidence level (sure, unsure, no idea). High confidence rates indicate reliable annotations: 91.9\% are "sure", 7.1\% are "unsure", and only 1.0\% are "no idea" (Figure \ref{fig:confidence}), effectively flagging ambiguous cases and potential non-local speakers. Of 6,907 annotated clips, 4,539 (65.7\%) were retained for modeling after manual review, with excluded segments predominantly containing Quranic recitation, music, or crosstalk.

Table~\ref{tab:agreement} reports inter-annotator agreement for each annotation category. The keep/skip decision achieves the highest raw agreement (91.76\%), with a Cohen's Kappa of 0.507 indicating moderate agreement, suggesting that the inclusion criteria were generally consistent despite the sensitivity of Kappa to class imbalance. The speech-type category also shows strong raw agreement (87.71\%) and a comparable moderate Kappa (0.586), reflecting occasional ambiguity in distinguishing single- and multi-speaker segments, particularly in talk-show formats with brief interjections.

The MSA/dialect distinction and emotion labels exhibit high percentage agreement (83.16\% and 90.53\%, respectively) but lower Kappa scores (0.310 and 0.179). This apparent discrepancy arises from the well-known sensitivity of Cohen's Kappa to class imbalance \cite{feinstein1990high}: when one category dominates (e.g., neutral emotion at 98\%, or the prevalence of dialectal speech), even moderate disagreement on minority classes substantially deflates Kappa while leaving raw agreement high. In terms of emotion, the overwhelming prevalence of neutral labels means that annotators rarely encountered non-neutral cases, limiting opportunities for both agreement and disagreement over minority emotions. For the MSA/dialect category, the challenge is compounded by the inherent difficulty of distinguishing MSA from dialect when speakers code-switch or produce regionally-accented MSA, a recognized source of annotator variability in Arabic speech corpora. Despite these moderate Kappa values, the high raw agreement rates and annotator confidence levels suggest that the labels are sufficiently reliable for training and evaluating dialect identification models.

\begin{table}[htbp]
\centering
\caption{Inter-Annotator Agreement Metrics}
\label{tab:agreement}
\begin{tabular}{lcc}
\toprule
\textbf{Annotation Category} & \textbf{Percentage Agreement} & \textbf{Cohen's Kappa} \\
\midrule
MSA or Dialect? & 83.16\% & 0.310 \\
Emotion & 90.53\% & 0.179 \\
Keep or Skip & 91.76\% & 0.507 \\
Type & 87.71\% & 0.586 \\
\bottomrule
\end{tabular}
\end{table}

\subsubsection{Audio quality}

\begin{figure}[h]
  \centering
  \includegraphics[width=0.95\textwidth]{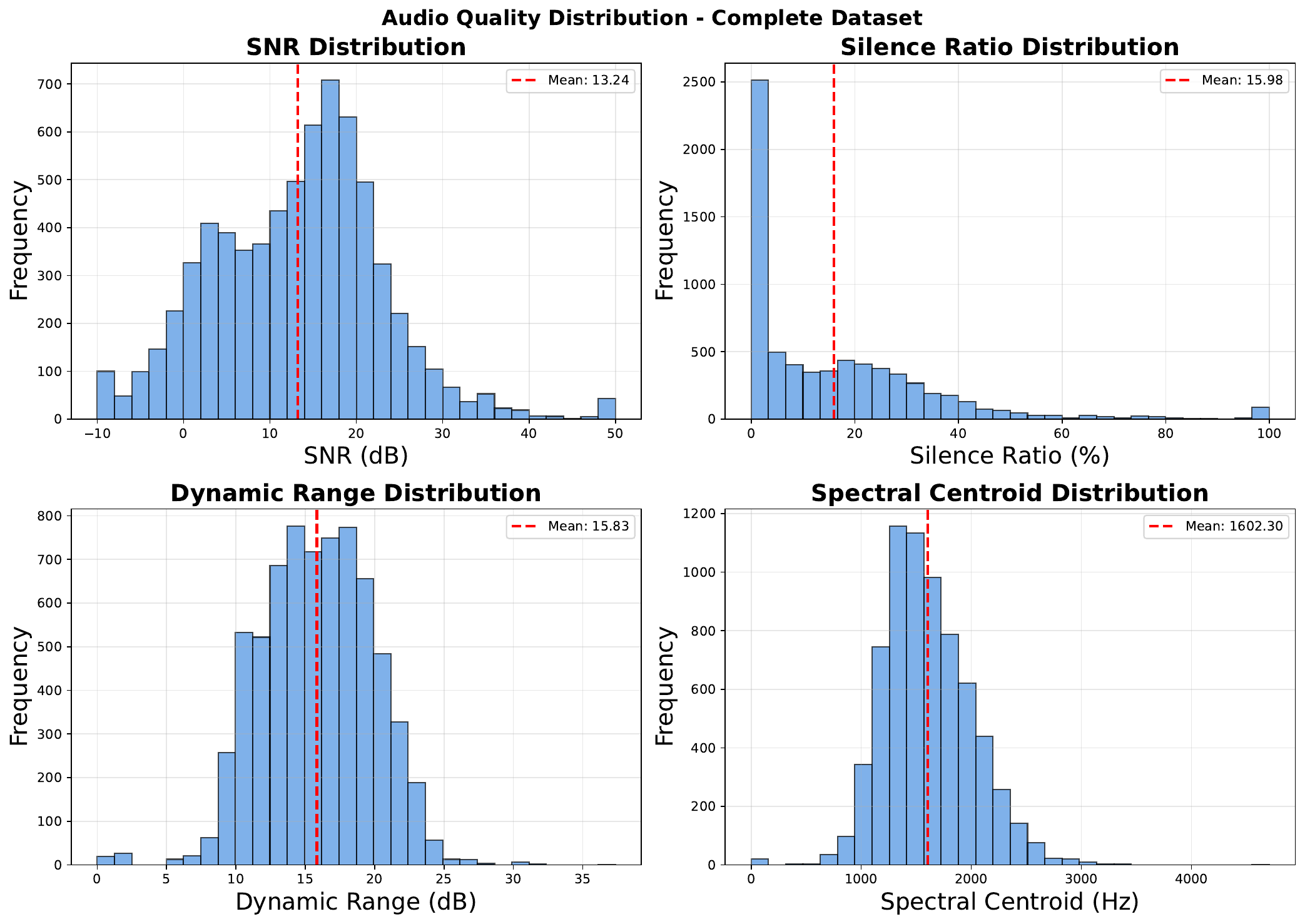}
  \caption{Distribution of audio quality metrics across the complete ARCADE dataset. The four panels show Signal-to-Noise Ratio (SNR), silence ratio, dynamic range, and spectral centroid. Red dashed lines indicate mean values.}
  \label{fig:audio_quality_full}
\end{figure}

To characterize the acoustic properties of the collected recordings, we computed four signal-level metrics for each clip: Signal-to-Noise Ratio (SNR), silence ratio, dynamic range, and spectral centroid. Figure~\ref{fig:audio_quality_full} presents the distribution of these metrics across the complete dataset.

The SNR distribution exhibits a roughly normal shape, centered at 13.24~dB, with values ranging from approximately $-10$~dB to over 50~dB. This range reflects the heterogeneous nature of radio broadcasts, which include studio-quality speech segments, field recordings, and phone-in programs with degraded audio. The silence ratio, defined as the percentage of frames below a speech activity threshold, shows a right-skewed distribution with a mean of 15.98\%, indicating that most clips contain predominantly active speech. The dynamic range averages 15.83~dB with a concentrated distribution between 10--20~dB, consistent with broadcast audio that undergoes dynamic range compression for transmission. The spectral centroid, a measure of the ``brightness'' of the audio signal, centers around 1602~Hz, typical of speech-dominated content.

\begin{figure}[h]
  \centering
  \includegraphics[width=0.95\textwidth]{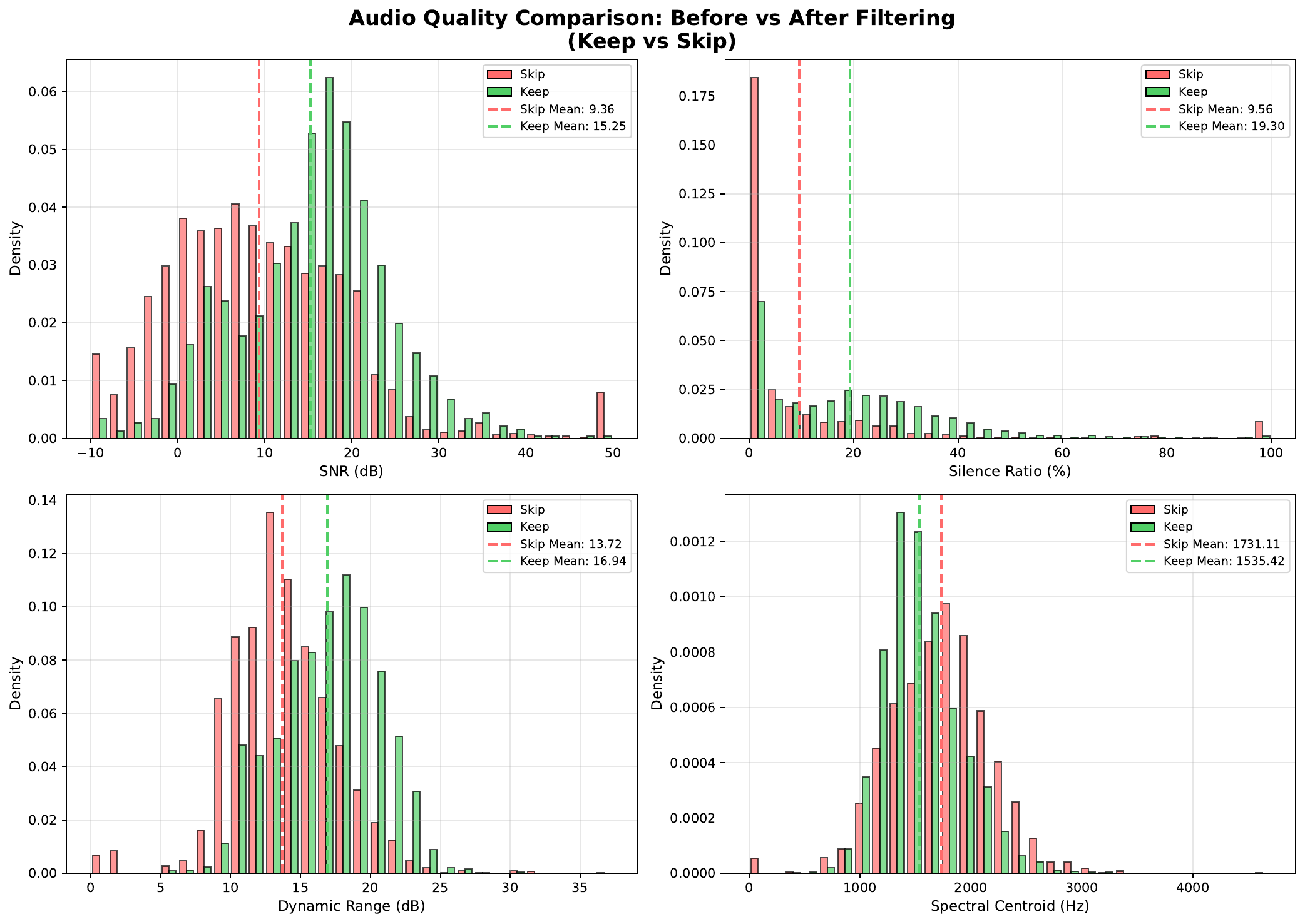}
  \caption{Comparison of audio quality metrics between retained ('Keep', green) and excluded ('Skip', red) segments. Systematic differences in SNR, silence ratio, dynamic range, and spectral centroid validate the manual filtering decisions.}
  \label{fig:audio_quality_keepskip}
\end{figure}

Figure~\ref{fig:audio_quality_keepskip} compares the audio quality distributions between retained (``Keep'') and excluded (``Skip'') segments. Notably, the annotation process did not explicitly consider audio quality; segments were excluded solely on the basis of content type (Quranic recitation, music, or crosstalk) to focus the corpus on dialectal speech. Nevertheless, the retained clips exhibit favorable acoustic properties as a byproduct of content-based filtering. Kept segments show higher mean SNR (15.25~dB vs.\ 9.36~dB for skipped clips), likely because music and Quranic recitation segments often include background instrumentation or reverberation that lowers SNR estimates. The silence ratio is higher for retained clips (19.30\% vs.\ 9.56\%), reflecting the natural pauses in conversational speech compared to continuous music. Dynamic range is also higher for kept segments (16.94~dB vs.\ 13.72~dB), and the spectral centroid is lower (1535~Hz vs.\ 1731~Hz), consistent with the exclusion of music, which typically exhibits broader high-frequency content than speech. These incidental quality improvements represent a positive outcome: the dialect-focused filtering yields a subset with above-average acoustic quality, which may benefit downstream modeling.

The geographic distribution of audio quality is visualized in Figure~\ref{fig:geo}, where circle colors encode the mean SNR per city on a gradient from red (low SNR, $\sim$0 dB) through yellow ($\sim$15 dB) to green (high SNR, $\sim$30 dB). Notable regional patterns emerge: North African cities, particularly those in Morocco and parts of Algeria, display predominantly green markers indicating high-quality recordings (mean SNR
> 25 dB). Similarly, Riyadh and Kuwait exhibit green markers, reflecting high-quality radio infrastructure in these Gulf capitals. In contrast, certain cities in Sudan and parts of the Levant exhibit red or orange markers, reflecting lower signal quality that may be attributable to streaming infrastructure, station equipment, or recording conditions. Other Gulf region cities show mixed quality, with UAE stations generally achieving moderate-to-high SNR values. These geographic variations in audio quality should be considered when training models, as they may confound acoustic degradation with dialectal features.

% \section{Dataset Analysis}

% In this section, we present descriptive statistics of the corpus to inform modeling decisions.  Figure~\ref{fig:dialect} shows that dialectal speech is twice as common as MSA, suggesting that models trained on this data may generalize better to dialects than formal MSA.  Figure~\ref{fig:types}
% reveals that regular single‑speaker recordings dominate, so we suggest that monologue speech may carry more distinct regional cues.  
% The cross‑tabulation in Figure~\ref{fig:crosstab} indicates that neutral
% emotion pervades all types, whereas positive emotions cluster in multi‑speaker shows, implying that emotional prosody may correlate with program type rather than location.

% The geographic plot (Figure~\ref{fig:geo}) shows clusters of recordings
% by country.  Many cities lie within a narrow latitude band (30–37°N),
% which may complicate learning latitude estimators but leaves more variation in
% longitude.  The top‑city bar chart (Figure~\ref{fig:topcountries}) highlights
% data imbalance: Doha and Manama supply nearly ten times more samples
% than smaller cities such as Alexandria or Ajman.  

\section{Discussion and Future Directions}

Our radio‑based corpus demonstrates that collecting geographically rich
Arabic speech can be obtained from public streaming services.  Annotating
emotion, speech type, and dialect categories allows multi‑task models
that jointly learn to predict location and other attributes.  However,
several challenges remain.  First, the data are imbalanced both
geographically and emotionally, with some cities over‑represented and
positive emotions under‑represented.  Future collection efforts should
target under‑sampled regions (e.g., rural areas) and programme types
that elicit varied emotions.  Second, the radio environment introduces
noise, music, and crosstalk; improved filtering or adaptive denoising
could enhance signal quality.
%  Third, gender balance remains an ongoing goal; although our guidelines encourage equal male and female representation, verifying gender requires automatic speaker diarisation or explicit annotation.

Looking ahead, the availability of fine-grained geographic metadata alongside linguistic annotations paves the way for novel research avenues. For instance, multi-task learning frameworks could jointly predict dialect category, emotion, and sub-regional origin, thereby disentangling phonetic cues from other latent factors. Furthermore, contrastive approaches that align audio representations with spatial identifiers offer a promising direction for future inquiry. We emphasize that while these modeling pursuits are beyond the scope of the present study, which focuses on data collection and descriptive analysis, we hope that releasing this corpus will catalyze broader interest in global audio geolocation and geolinguistics.

\section{Data Availability}

The complete ARCADE dataset is openly available on the Hugging Face Datasets platform\footnote{\url{https://huggingface.co/datasets/riotu-lab/ARCADE-full}}. All data files are distributed under an open research license (CC BY 4.0) and are intended for non-commercial academic and research use. The repository includes the full audio corpus, annotation files with emotion, speech type, dialect category, and quality labels, detailed documentation, and metadata linking each clip to its source city and country. The structured release supports the FAIR data principles (Findable, Accessible, Interoperable, and Reusable) and is permanently archived for reproducible research.

\section{Conclusion}

We introduced ARCADE, a novel corpus of Arabic radio speech annotated with city-level geographic labels and rich metadata, addressing a critical gap in fine-grained dialect identification resources. Unlike existing corpora that provide country or region-level annotations, ARCADE offers recordings from 58 cities spanning 19 Arab countries, enabling research on sub-regional dialectal variation that was previously infeasible due to data limitations.

Our collection pipeline leverages publicly available radio streams from platforms such as Radio Garden and World Radio Map, demonstrating that large-scale, geographically diverse speech data can be harvested from existing broadcasting infrastructure without requiring costly field recordings. The pipeline ensures at least ten recordings per city, yielding 3,790 unique 30-second segments. Each clip is annotated with emotion, speech type, dialect category, and annotator confidence, supporting multi-task learning approaches that jointly leverage these complementary labels.

The descriptive analysis reveals several patterns with implications for downstream modeling. Neutral emotion dominates the corpus, reflecting the professional register of broadcast speech, while dialectal content constitutes the largest category. The prevalence of single-speaker monologue segments suggests that radio speech may carry distinct regional cues amenable to dialect identification. However, pronounced geographic imbalance across cities will necessitate sampling strategies or weighted loss functions to ensure equitable model performance. Audio quality analysis further reveals systematic geographic variation in SNR, with implications for disentangling acoustic degradation from dialectal features.
By releasing this corpus alongside transparent documentation of our collection and annotation protocols, we provide the research community with a foundation for advancing fine-grained dialect attribution from audio. Potential applications extend beyond dialect identification to sociolinguistic studies, robustness evaluation under domain shift, and transfer learning to related media such as telephony and online video. The reusable pipeline and annotation guidelines also facilitate future expansions and community contributions. Detailed experimental benchmarking is left to subsequent studies.

\bibliographystyle{unsrt}
\bibliography{references}

\end{document}